\pdfoutput=1

\documentclass[11pt]{article}

\usepackage[]{acl}

\usepackage{times}
\usepackage{latexsym}

\usepackage[T1]{fontenc}

\usepackage[utf8]{inputenc}

\usepackage{microtype}
\usepackage{booktabs}
\usepackage{graphicx}
\usepackage{amsthm}
\usepackage{amsmath}
\usepackage{amsfonts}
\usepackage{xspace}
\usepackage[]{algorithm2e}
\usepackage{bm}
\usepackage{pifont}
\usepackage{multirow}

\def\smallcol{\hskip 6pt}
\def\tinycol{\hskip 2pt}

\definecolor{cadmiumgreen}{rgb}{0.0, 0.42, 0.24}
\definecolor{cerulean}{rgb}{0.0, 0.48, 0.65}
\definecolor{cadmiumorange}{rgb}{0.93, 0.53, 0.18}

\def\ours{OmniTab\xspace}
\def\wtq{WikiTableQuestions\xspace}
\def\squall{SQUALL\xspace}
\def\wikisql{WikiSQL\xspace}

\title{\ours: Pretraining with Natural and Synthetic Data \\ for Few-shot Table-based Question Answering}

\author{Zhengbao Jiang$^{1}$\thanks{~~Work was done when interning at Microsoft Azure AI.}, Yi Mao$^2$, Pengcheng He$^2$, Graham Neubig$^1$, Weizhu Chen$^2$ \\
$^1$Language Technologies Institute, Carnegie Mellon University \quad$^2$Microsoft Azure AI \\
\texttt{\{zhengbaj,gneubig\}@cs.cmu.edu} \\
\texttt{\{maoyi,penhe,wzchen\}@microsoft.com}}

\begin{document}
\maketitle
\begin{abstract}
The information in tables can be an important complement to text, making table-based question answering (QA) systems of great value.
The intrinsic complexity of handling tables often adds an extra burden to both model design and data annotation.
In this paper, we aim to develop a simple table-based QA model with minimal annotation effort.
Motivated by the fact that table-based QA requires both alignment between questions and tables and the ability to perform complicated reasoning over multiple table elements, we propose an omnivorous pretraining approach that consumes both \emph{natural} and \emph{synthetic} data to endow models with these respective abilities.
Specifically, given freely available tables, we leverage retrieval to pair them with relevant natural sentences for mask-based pretraining, and synthesize NL questions by converting SQL sampled from tables for pretraining with a QA loss.
We perform extensive experiments in both few-shot and full settings, and the results clearly demonstrate the superiority of our model \ours, with the best multitasking approach achieving an absolute gain of 16.2\% and 2.7\% in 128-shot and full settings respectively, also establishing a new state-of-the-art on \wtq.
Detailed ablations and analyses reveal different characteristics of natural and synthetic data, shedding light on future directions in omnivorous pretraining.%
\footnote{Code, pretraining data, and pretrained models are available at \url{https://github.com/jzbjyb/OmniTab}.}
\end{abstract}

\section{Introduction}\label{sec:intro}
\begin{figure*}[tb]
\begin{minipage}{\textwidth}
\begin{minipage}[b]{0.61\textwidth}
\centering
\includegraphics[width=\columnwidth, clip, keepaspectratio]{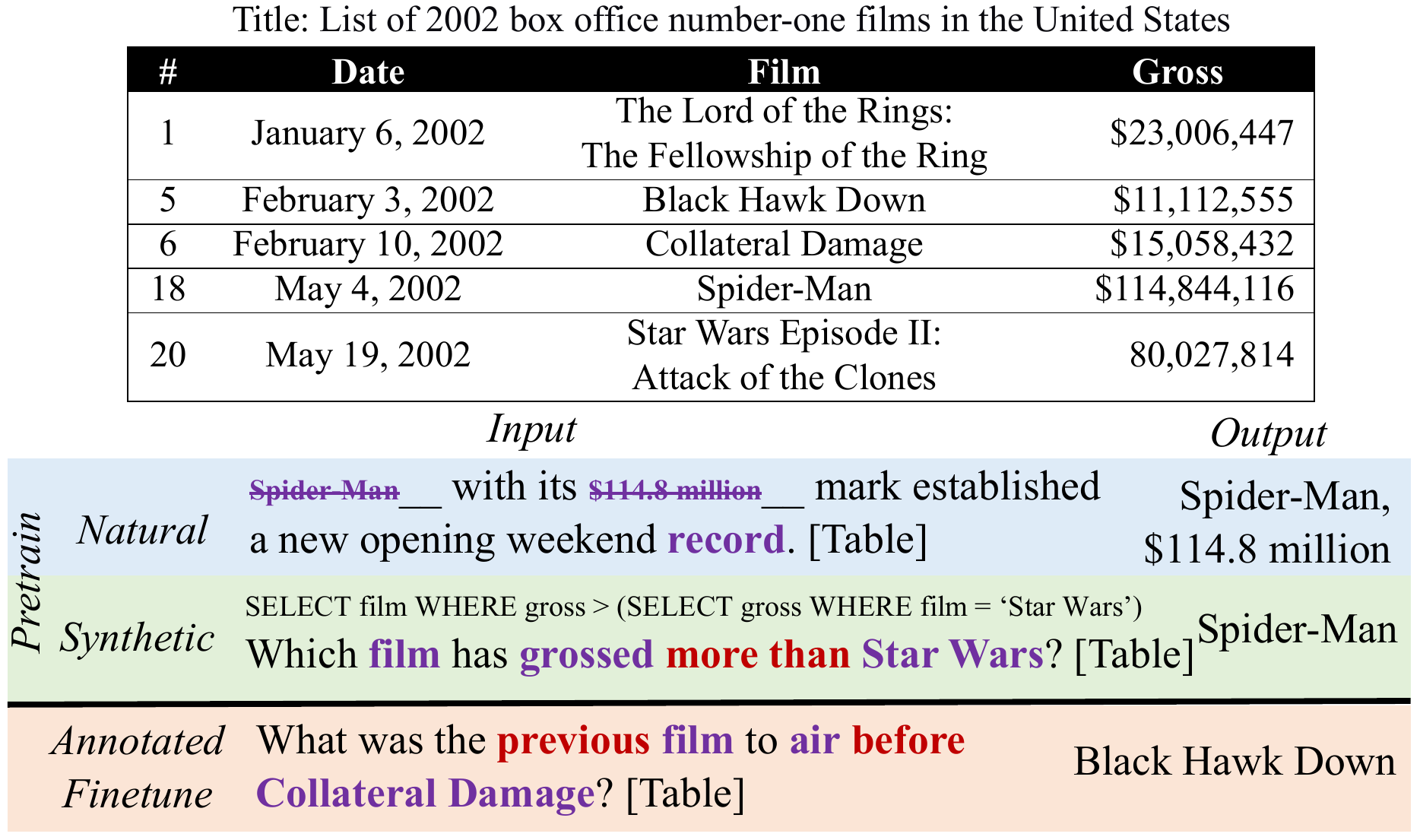}
\captionof{figure}{Example of natural and synthetic pretraining data and a manually annotated finetuning question. Phrases aligned with table elements and reasoning operations are marked with \textcolor{violet}{violet} and \textcolor{red}{red} respectively. [Table] denotes the linearized table.}
\label{fig:example}
\end{minipage}
\hfill
\begin{minipage}[b]{0.37\textwidth}
\centering
\includegraphics[width=\columnwidth, clip, keepaspectratio]{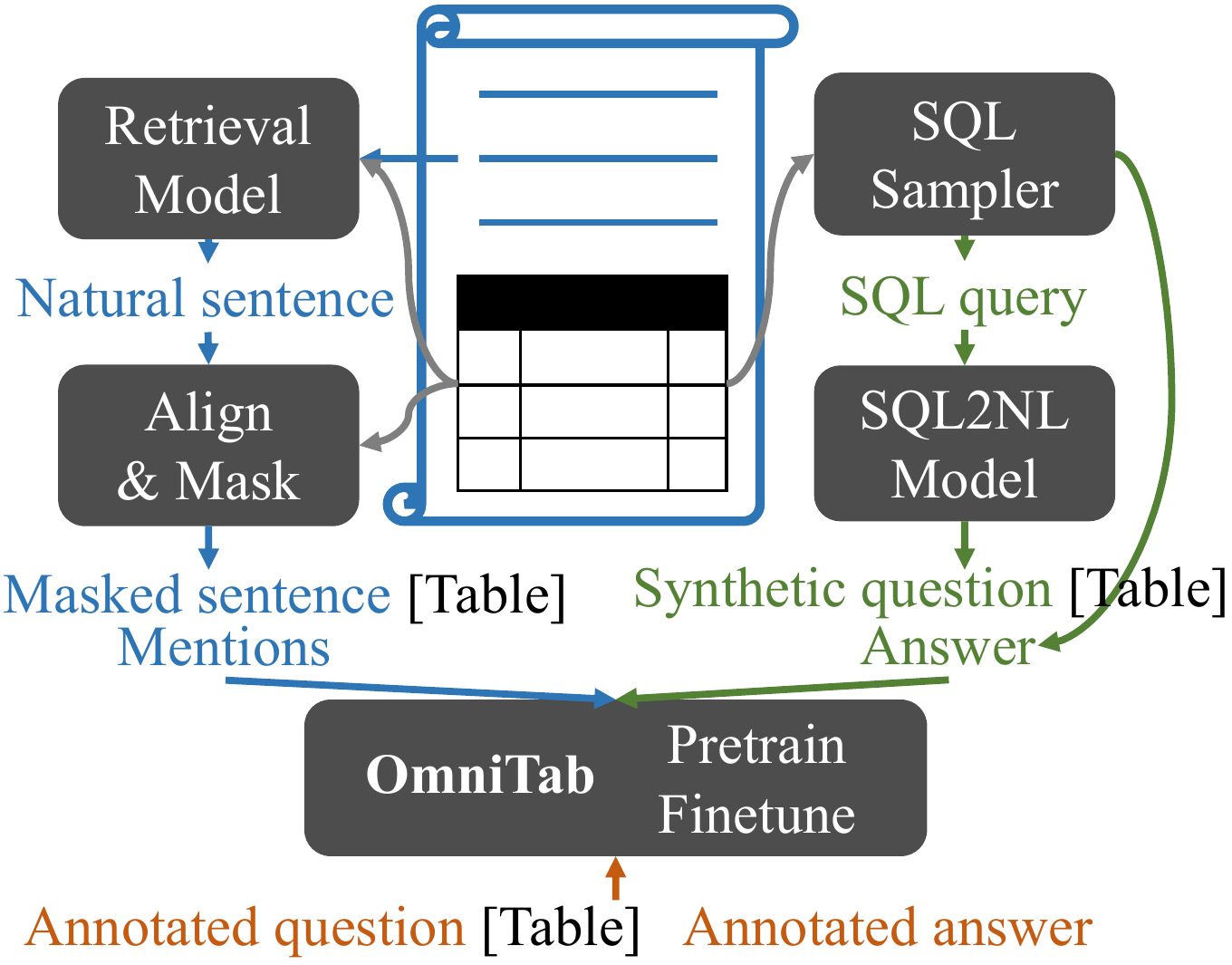}
\captionof{figure}{The overall framework of generating and pretraining with natural data (\textcolor{cerulean}{blue} pipeline), synthetic data (\textcolor{cadmiumgreen}{green} pipeline), and finetuning with limited annotated questions (\textcolor{cadmiumorange}{orange} pipeline) for our \ours model.}
\label{fig:framework}
\end{minipage}
\end{minipage}
\end{figure*}

Humans are voracious information omnivores, consuming information in a number of forms.
Unstructured text is the form covered in most work in NLP, but another form widely used on the web, academic papers, and reports is the \emph{table}, where elements are organized in rows and columns and presented in a structured and succinct way.
Because of this, systems to aid information access over tables, such as table-based question answering (QA) \cite{wtq-2015-pasupat,sqa-2017-iyyer,wikisql-2017-zhong}, hold significant promise.
However, they also require understanding of the table structure and sophisticated reasoning across multiple elements to get the final answer.
This intrinsic complexity not only often requires special-purpose model designs such as pipeline systems that generate structured queries as intermediate steps \cite{liang-2018-mapo,wang-2019-latent,tabert-2020-yin,yu-grappa-2021}, but also adds an extra burden to the process of data annotation \cite{wtq-2015-pasupat,shi-2020-squall}.

Given the challenges above, we ask the question: ``can we create a \emph{simple} model that is able to answer complex table-based questions with \emph{minimal annotation effort}?''
Both modeling simplicity and limited assumptions regarding data availability would make it easier to apply table-based QA systems in practical scenarios.
To this end, we focus on developing a simple and generic end2end table-based QA model where only a limited number of annotated natural language (NL) questions are available; the first attempt to address this problem under few-shot settings.
In order to answer table-related questions, an end2end model needs to understand both the NL question and the table, build connections between the two formats, then perform complex reasoning.
Taking the manually annotated question in \autoref{fig:example} as an example, models need to align entities (``Collateral Damage'') and concepts (``film'', ``air'') in the question to elements in the table (the ``Collateral Damage'' cell and the ``Film'' and ``Date'' columns) and perform comparative reasoning based on chronological order (indicated by ``previous'' and ``before'') to find the final answer.
Motivated by this, we propose \textbf{\ours}, an omnivorous pretraining approach that consumes \emph{natural} data to endow models with the ability to understand and align NL with tables, and \emph{synthetic} data to train models to perform reasoning.

To obtain natural NL-table pairs, we propose a novel approach that leverages the multitude of tables freely available from the web, and uses \emph{retrieval} to pair them with relevant NL sentences.
Compared with manually defined heuristics used in previous work \cite{tapas-2020-herzig,tabert-2020-yin}, retrieval-based methods have the potential to discover better-aligned sentences.
We explore different retrieval methods including string-based matching, sparse retrieval, and dense retrieval \cite{dpr-2020-karpukhin,khattab-2020-colbert,gao-2021-coil}.
Given these retrieved pairs, phrases in the sentence that align with table elements are then masked and the model takes both the masked sentence and the linearized table as input to predict masked mentions.
For example, in \autoref{fig:example} the retrieved sentence describes a particular row and contains two phrases matching cells in the table (i.e., ``Spider-Man'' and ``\$114.8 million'') which are masked for prediction.
To perform this sort of masked prediction, models need to understand that the sentence is about a record-breaking movie and refer to the table to extract the correct cells.
Thus, training on this data endows models with better understanding and alignment across both formats.

For the synthetic data approach, we propose a method where structured queries such as SQL are first sampled and then converted into NL questions using an SQL2NL model, which allows for control of the reasoning operations covered by the SQL.
Compared to existing work that trains directly on SQL \cite{tapex-2021-liu}, an approach hindered by the gap between structured and natural language, training on synthetic NL questions can close the gap, especially when limited annotated NL questions are available.
We train the SQL2NL model with a small number of SQL-NL pairs and further boost its performance using verification-based self-training, which selects high-quality generated NL questions based on their likelihood to generate the gold answer.
The converted NL question concatenated with the linearized table is fed into the model to directly generate the final answer, as shown in the synthetic example in \autoref{fig:example} which involves comparative reasoning indicated by the phrase ``more than''.
Although the fluency and naturalness of synthetic data is usually lower than natural data, learning on synthetic data provides a direct way to simulate different reasoning skills, which is relatively sparse in natural data.

Our overall framework is shown in \autoref{fig:framework}.
We use tables from Wikipedia and retrieve relevant sentences from the same page to generate natural text-table parallel data after masking mentions aligned to table elements (the blue pipeline on the left of \autoref{fig:framework}).
We use SQL queries sampled by \citet{tapex-2021-liu} and convert them to NL questions as synthetic text-table parallel data (the green pipeline on the right of \autoref{fig:framework}).
We use \wtq (WTQ) \cite{wtq-2015-pasupat}, a widely used table-based QA dataset consisting of complex questions, as our major benchmark to evaluate our pretraining methods, and further use \wikisql \cite{wikisql-2017-zhong} and topic-categorized WTQ \cite{chemmengath-2021-wtqtopic} to evaluate the robustness of our methods, all under few-shot setting with sizes ranging from 16 to 1,024.
When using only 128 annotated questions, our model \ours improves over the best-performing baseline by an absolute gain of 13.2\% and 12.3\% with natural and synthetic pretraining separately and 16.2\% when combined, demonstrating the effectiveness of the approach.
We also achieve state-of-the-art performance on the full WTQ with an absolute gain of 2.7\%.
Extensive ablations and analyses reveal that natural and synthetic data indeed play the role of enhancing alignment and injecting reasoning, shedding light on future works on omnivorous pretraining.

\section{End2End Table-based QA}\label{sec:e2e_tableqa}
In this section, we first explain the setting of table-based QA, then introduce the input format as well as our model architecture.

\paragraph{Table-based QA}
Each example in table-based QA consists of an NL question $\bm{q}$, a table $T$, and an answer $\bm{a}$.
Both questions and answers are a sequence of tokens.
Each table consists of $N$ rows $\{r_i\}_{i=1}^{N}$ and $M$ columns, where the token sequence in the cell located at the $i$-th row and $j$-th column is denoted as $\bm{c}_{i,j}$.
The first row $r_1$ is the header row, indicating the meaning of each column.
Table-based QA aims to generate the answer $\bm{a}$ given both the question $\bm{q}$ and the table $T$ as the input.

\paragraph{Input Format}
We follow \citet{tapex-2021-liu} in concatenating the NL context with the linearized table as input.
We flatten the table following a top-to-bottom and left-to-right order, where we prepend ``col:'' to the beginning of the header and ``row $i$:'' to the beginning of the $i$-th row to separate different rows: $T$ = [col: $r_1$ row 1: $r_2$ ... row $N$: $r_N$].
Cells within each row are separated by a vertical bar ``|'' $r_i$ = [$\bm{c}_{i,1}$ | $\bm{c}_{i,2}$ | ... | $\bm{c}_{i,M}$].
Finally, the question $\bm{q}$ is prepended to the linearized table: [$\bm{q}$ $T$].

\paragraph{Model Architecture}
We use the state-of-the-art table-based QA model TAPEX \cite{tapex-2021-liu} as our base model, which is based on BART \cite{lewis-2020-bart}. It feeds questions and tables into the encoder and generates answers from the decoder: $P(\bm{a}|\bm{q}, T)$. 
Multiple answers are joined with commas into a single output sequence.

\section{\ours: Pretraining with Natural and Synthetic Data}
As mentioned in the introduction, table-based QA requires both (1) the ability to align NL phrases with table elements that could be expressed in different wording and (2) perform reasoning such as filtering, aggregation, superlatives, comparatives, and arithmetic.
Compared to synthetic data, real NL sentences relevant to the table excel at enhancing the first capability since they are more natural and fluent, exhibiting various language variations. 
Learning on real sentences can endow models to grasp the nuances in language and align with structured tables.
On the other hand, synthetic data is flexible, manipulable, and easy to obtain.  
It is costless to generate synthetic data via manipulating different aspects of the generated data to incorporate various desired properties.
As a result, we can generate large amounts of complicated synthetic data covering various reasoning operations, which is lacking in the NL corpora. 
This motivates us to explore both types of data.

\subsection{NL-Table Alignment Through Retrieval}\label{sec:natural}
Using the Wikipedia table corpus released by \citet{tapas-2020-herzig}, we explore three retrieval methods to construct aligned NL-Table pairs and propose a new pretraining objective.

\subsubsection{Retrieval Protocol}

Since sentences relevant to a table are usually included in the same document, we restrict our retrieval models to only consider the document containing the table, with the purpose of reducing noise and increasing efficiency.

\paragraph{String-based Matching}
For each sentence, we enumerate over all cells in the table and find the longest common substring (LCS) between a cell $\bm{c}_{i,j}$ and a sentence $\bm{s}$.
An LCS is considered as a mention to be masked if it (1) is not a stopword, (2) contains alphanumeric characters, (3) is a complete word, and (4) is longer than 70\% of the length of the cell.
We choose the sentence with the largest number of mentions to pair with the table.

\paragraph{Sparse Retrieval with BM25}
Another method of string-based matching is BM25 with tables as queries and sentences as candidates. 
Different from the above method matching whole cells, BM25 treats tables as a bag of tokens.
We linearize tables as queries and choose the most relevant sentence to compose the pair.
Since BM25 retrieval does not return aligned phrases and cells, we resort to the above method detect mentions.

\paragraph{Dense Retrieval with Token Representations}
The above two methods can only consider exact string matches, which is sub-optimal because different expressions might be used between sentences and tables such as ``\$114,844,116'' and ``\$114.8 million'' in \autoref{fig:example}.
Tables tend to use full and formal expressions, while expressions in NL tend to be casual, often with abbreviations.
To address this issue, we propose to use dense representations to perform soft matching.
Many works use a single dense vector to represent the whole query/document for retrieval \cite{guu-2020-realm,dpr-2020-karpukhin}.
However, in our case, queries are tables usually consisting of many cells,\footnote{Wiki tables have 26 cells on avg with extreme cases 500+.} thus representing a whole table as a single vector might lead to information loss, which performs well below BM25 in our preliminary experiments.
Additionally, retrieval based on a single vector only returns sentence-table pairs without revealing phrase-cell alignment, which is required for masking purposes.
Thus, we propose to use token representation to directly match phrases and cells, similar to token-level dense retrieval \cite{khattab-2020-colbert,gao-2021-coil} in spirit.

\begin{figure}
\centering
\includegraphics[width=\columnwidth, clip, keepaspectratio]{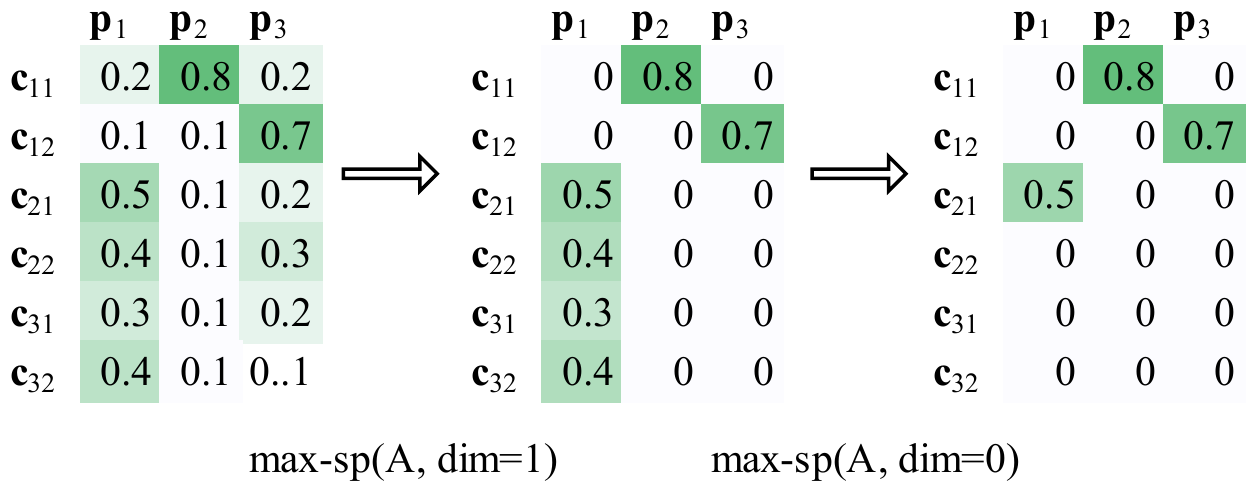}
\caption{Applying max-sparsify on a relevance matrix of cells and phrases on two dimensions consecutively.}
\label{fig:dense}
\end{figure}

Specifically, we use BART to obtain token representations for each sentence $\bm{s}$ and table $T$ separately.
We then use a named entity recognition (NER) model to detect candidate phrases $\{\bm{p}_l\}_{l=1}^{L}$ in the sentence.
Each phrase and cell are represented as the average token representation, denoted as $\bm{e}_{\bm{p}_l}$ and $\bm{e}_{\bm{c}_{i,j}}$ respectively after normalized to a unit vector.
We compute a similarity for each cell-phrase pair using dot product $\bm{e}_{\bm{c}_{i,j}} \cdot \bm{e}_{\bm{p}_l}$, resulting in a two-dimensional similarity matrix $A \in \mathbb{R}^{NM \times L}$, where each row and column correspond to a cell and a phrase respectively.
We aggregate the relevance matrix $A$ to derive relevance scores for ranking sentences and an assessment score for each phrase to choose salient mentions for masking.
Given the fact that soft match based on dense vectors usually yields a non-zero relevance score even for irrelevant pairs, we apply the max-sparsify operator to emphasize relevant matches and eliminate noisy irrelevant matches, similarly to the max operator in \citet{khattab-2020-colbert,gao-2021-coil}.
The $\text{max-sp}(A, \text{dim}=i)$ keeps the max entry along dimension $i$ of the matrix $A$ and changes other entries to zero.
As illustrated in \autoref{fig:dense}, we first apply this operator over all phrases ($\text{dim}=1$), assigning each cell the best-matching phrase, then apply it over all cells ($\text{dim}=0$), assigning each remaining phrase to its best-matching cell.
We use the sum of the sparsified matrix as the relevance score to choose the best-matching sentences, rank remaining phrases in that sentence ($\bm{p}_2>\bm{p}_3>\bm{p}_1$ in \autoref{fig:dense}), and choose phrases with scores higher than a threshold $\tau$ as mentions to be masked.
\begin{equation}
\text{rel}(\bm{s}, T)=\text{sum}(\text{max-sp}(\text{max-sp}(A, 1), 0)).
\end{equation}

\subsubsection{Learning Objective}

Given a retrieved sentence $\bm{s}$ associated with the table $T$, we apply two masking strategies: (1) randomly mask tokens in the sentence or cells in the table (2) salient mention masking where we first identify phrases in the sentence that align with table elements, then mask aligned phrases (denoted as mentions).
Compared to random masking, salient masking specifically focuses on masking shared information, enabling the model to better learn the correspondence across formats, which we will verify in \autoref{sec:exp_ablation}.
Since we use TAPEX as the base model, which is based on BART, we follow the pretraining format of BART to generate the original unmasked sequence given the input with masked tokens (in either NL or table).
Instead of computing the negative log likelihood loss (NLL) over all tokens, we only compute at masked positions to emphasize the recovery of missing information:
\begin{equation}
\mathcal{L}_{\text{mask}} = - \log P_{\text{mask}}(\bm{s}, T|\bm{s}^*, T^*),\label{eq:natural}
\end{equation}
where $\bm{s}^*$/$T^*$ denote the masked sentence/table, and $P_{\text{mask}}(\cdot|\cdot)$ only computes loss at masked positions.

\subsection{Synthetic Questions Converted from SQL}\label{sec:synthetic}
We use synthetic questions to simulate real table-based questions that involve various reasoning operations, such as the comparative operation in \autoref{fig:example}.
While directly synthesizing complex NL questions is challenging, it is easier to generate complex structured queries such as SQL because they follow predefined syntax, and different reasoning operations can be implemented with different SQL templates.
For example, the SQL query in \autoref{fig:example} is based on a template that compares entries w.r.t.~a numerical property.
This motivates us to first generate SQL ($\bm{o}$) then convert them to NL questions ($\bm{q}$).

Fortunately, \citet{tapex-2021-liu} already sampled a large corpus of SQL with associated answers based on tables from Wikipedia and used SQL plus tables as input to pretrain their model TAPEX.
They achieved state-of-the-art performance on table-based QA, making TAPEX a strong base model.
However, there is a large gap between SQL and NL questions, and training solely on SQL might hinder it from closing this gap.
Instead, we use NL questions in the pretraining directly.
Given synthesized NL questions, we pretrain with a standard generative QA loss that takes NL questions concatenated with tables as input and decode answers obtained by executing SQL queries over tables:
\begin{equation}
\mathcal{L}_{\text{QA}} = - \log P(\bm{a}|\bm{q}, T).\label{eq:synthetic}
\end{equation}

\begin{figure}
\centering
\includegraphics[width=0.85\columnwidth, clip, keepaspectratio]{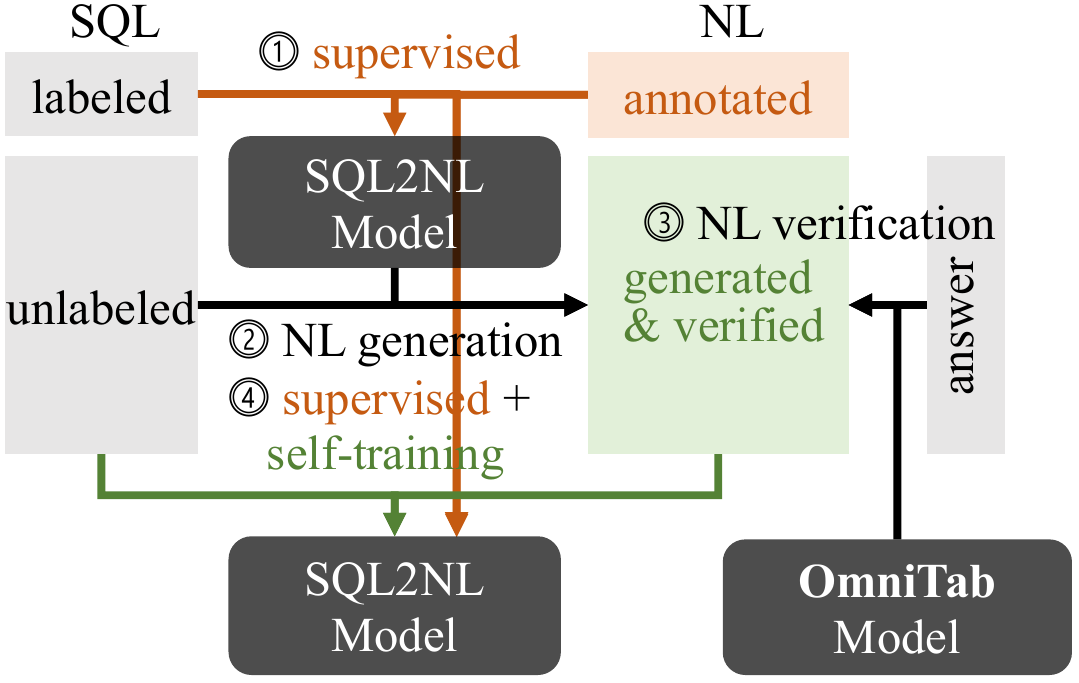}
\caption{The procedure of learning a SQL2NL model with verification-based self-training. Annotated / generated NL questions are denoted as \textcolor{cadmiumorange}{orange} / \textcolor{cadmiumgreen}{green}.}
\label{fig:selftraining}
\end{figure}

\paragraph{Sampling SQL Based on Templates}
\citet{tapex-2021-liu} leverage tables from WTQ and instantiate different templates over them to sample large amounts of SQL, where answers are obtained by execution.
Templates are automatically extracted from \squall \cite{shi-2020-squall}, which includes SQL corresponding to NL questions in WTQ.

\paragraph{Training SQL2NL Models}
We use BART as our base model and finetune it with limited SQL-NL pairs to strictly conform to the few-shot setting.
We use \squall to simulate few-shot scenarios, by assuming that only a limited number of SQL queries have annotated NL questions, which we elaborate in \autoref{sec:exp_setting}.
The model takes SQL as input and generates NL questions autoregressively.%
\footnote{We tried adding the corresponding table as an additional input but found no gain despite increased computational cost.}

\paragraph{Self-training with Verification-based Selection}
Even with a strong model like BART, the accuracy of SQL2NL is not perfect, especially in the face of limited data.
To further improve performance, we devise a verification-based self-training approach that selects high-quality NL questions generated from unlabeled SQL by assessing how likely they elicit correct answers from the table-QA model.

As illustrated in \autoref{fig:selftraining}, we first finetune BART model on supervised SQL-NL pairs to obtain the initial SQL2NL model (\ding{172}), which is used to generate NL questions for unlabeled SQL (\ding{173}) in the second step.
We attempted two generation methods including beam search and top-k sampling \cite{fan-2018-hierarchy} and found that beam search leads to more diverse results. Thus we use a beam size of 50 to generate candidate NL questions $\hat{\bm{q}}$.
Third, we choose high-quality candidates for self-training based on various criteria.
The most straightforward criterion is to choose questions with the highest generation probabilities for self-training $\text{score}_{\text{gen.}}(\hat{\bm{q}})=P^{\text{SQL2NL}}(\hat{\bm{q}}|\bm{o})$, which does not lead to improvement as we will show in the ablations (\autoref{sec:exp_ablation}).
Motivated by the fact that \ours has a strong capacity to answer table-related questions after large-scale pretraining and finetuning, we rank generated sentences by verifying how likely they elicit the correct answer using \ours $\text{score}_{\text{ver.}}(\hat{\bm{q}})=P(\bm{a}|\hat{\bm{q}},T)$, which provides a simple and effective way to leverage the QA capacity of \ours to indirectly provide feedback to the SQL2NL model (\ding{174}).
Given the verification scores, we first pair each unlabeled SQL with the sentence with the highest score among 50 candidates, then keep the top-$K$ SQL-NL ranked based on the score as our self-training data.
At the last step, the small supervised data is combined with the enlarged self-training data to train the final SQL2NL model (\ding{175}).

\subsection{Combining Natural and Synthetic Data}\label{sec:multitask}
We perform multitasking using the two previously defined objectives (\autoref{eq:natural}, \autoref{eq:synthetic}) to combine natural and synthetic data.
In addition, since we use TAPEX as initialization and their SQL-based pretraining has demonstrated effective in endowing models with reasoning capability, we add SQL-based pretraining in the multitask setting.
$\mathcal{L}_{\text{sql}}=-\log P(\bm{a}|\bm{o},T)$, resulting a combination of three parts $\mathcal{L}_{\text{mask}}+\mathcal{L}_{\text{QA}}+\mathcal{L}_{\text{sql}}$ as the multitask loss.

\section{Experiments}
We first detail the experimental settings (\autoref{sec:exp_setting}).
Then we report on extensive experiments, starting with the overall result including all design elements to achieve the best results (\autoref{sec:exp_overall}), then breaking down into each design choice to analyze their individual contribution (\autoref{sec:exp_ablation}).
Finally, we analyze concrete examples to provide insight on different characteristics of natural and synthetic data (\autoref{sec:exp_analysis}).

\subsection{Experimental Settings}\label{sec:exp_setting}
\paragraph{Few-shot Settings}
We use \wtq \cite{wtq-2015-pasupat} as our major benchmark, as it is a widely used table-based QA dataset involving various complex reasoning operations, and report the answer accuracy given by the official evaluator.
Following work on few-shot QA \cite{ram-2021-fewshotqa}, we create few-shot settings of WTQ by sampling a subset from the original training containing 11K examples in total, with sizes changing on a logarithmic scale from 16 to 1024.

Another component that requires annotated NL questions is the SQL2NL model.
We use \squall \cite{shi-2020-squall}, which contains $\approx$10K annotations in total, to simulate few-shot scenarios by varying the number of SQL annotated with NL from 16 to 4,096.
In the $f$-shot setting, we use \squall-$f$ to train the SQL2NL model and WTQ-$f$ to finetune QA models.
Since \squall and WTQ share the same set of NL questions, we make sure that \squall-$f$ also includes the same questions as WTQ-$f$, so in total only $f$ annotated questions are used in the $f$-shot setting.  

We also report on WikiSQL \cite{wikisql-2017-zhong}, another table-based QA benchmark with relatively simple questions.
To evaluate robustness under topical shift, we further use \wtq-TS which split WTQ into five topics \cite{chemmengath-2021-wtqtopic} based on Wikipedia categories.
We follow their creation procedure to reproduce the split, and evaluate our methods by finetuning on one topic and testing on the other four topics.

\paragraph{Pretraining Corpora}
We use the table corpus by \citet{tapas-2020-herzig} extracted from Wikipedia as our source of tables for retrieval.
All tables are preprocessed into a two-dimensional structure with a single header and one or multiple data rows.
We use a subset of this corpus and find the corresponding Wikipedia page through its URL, which is preprocessed into sentences using SLING.
Since some tables are noisy and some Wikipedia pages do not contain meaningful sentences, eventually we pair approximately 0.5M tables with sentences using our three retrieval methods.
To make the synthetic data of similar size, we also use 0.5M SQL sampled by \citet{tapex-2021-liu} to generate synthetic questions.

\paragraph{Baselines}
We consider two types of models as baselines (1) pipeline methods that execute generated SQL to get answers such as \textbf{TaBERT} \cite{tabert-2020-yin} with \textbf{MAPO} \cite{liang-2018-mapo} as the semantic parser and (2) end2end methods that directly generate answers, such as \textbf{BART} \cite{lewis-2020-bart}, \textbf{TAPAS} \cite{tapas-2020-herzig}, and \textbf{TAPEX} \cite{tapex-2021-liu}.
More discussions about table-related models can be found in the related work section in \autoref{sec:related}.
Since we also incorporate the SQL data used by TAPEX in our final multitask setting, we report \textbf{TAPEX$^*$} when comparing with our multitask setting, which continued to train TAPEX on SQL data for as many steps as \ours to make a fair and rigorous comparison.
We use \textbf{\ours} to denote our full model trained in the multitask setting with both natural, synthetic, and SQL data (\autoref{sec:multitask}).

\paragraph{Implementation Details}\label{sec:implementation}
During pretraining, we use a batch size of 512 and train \ours for 5 epochs, which takes about 20 hours on 8 V100 GPUs for multitasking on both natural and synthetic data.
During finetuning, we use a batch size of 96 and finetune \ours for 50 epochs, which takes about 30 minutes on 8 V100 GPUs.
We use a learning rate of 2e-5 for both pretraining, finetuning.
We use BART-large and TAPEX-large in all experiments.
For dense retrieval, since we use the max operation, all phrases have scores $\in [-1, 1]$.
We bucket phrases into bins with an interval of 0.1, manually inspect the quality of some randomly sampled phrases from each bin, and found that phrases with scores larger than $\tau=0.6$ are of high quality.
We use spaCy\footnote{\url{https://spacy.io/}} to detect named entities for dense retrieval.
For self-training of the SQL2NL model, we use the best-performing \ours model without self-training as the verification QA model, and make sure that it uses the same amount of annotations as the final model (i.e. if the final model is a $f$-shot model, we also use $f$ annotations to train the verification model).
In our final model, we added approximately 10K SQL-NL pairs for self-training.

\begin{table}[tb]
\small
\centering
\begin{tabular}{@{}lr@{\smallcol}r@{\smallcol}r@{\smallcol}r@{}}
\toprule
\textbf{Model \hfill{$f$-shot:}} & \textbf{16} & \textbf{128} & \textbf{1024} & \textbf{full} \\
\midrule
\multicolumn{5}{c}{\emph{Pipeline systems}} \\
TaBERT+MAPO \cite{tabert-2020-yin} & 7.7 & 15.1 & 33.3 & 52.3 \\
\midrule
\multicolumn{5}{c}{\emph{End2end systems}} \\
BART \cite{lewis-2020-bart} & 2.9 & 8.4 & 17.3 & 38.0 \\
TAPAS \cite{tapas-2020-herzig} & 9.8 & 18.9 & 33.6 & 48.8 \\
TAPEX \cite{tapex-2021-liu} & 10.4 & 23.1 & \emph{45.5} & 59.5 \\
TAPEX$^*$ & \emph{15.7} & \emph{25.2} & 44.6 & \emph{60.1} \\
\midrule
\multicolumn{5}{c}{\emph{Ours (end2end)}} \\
\ours w/ natural data & 22.8 & 38.4 & 49.8 & 61.3 \\
\ours w/ synthetic data & 21.5 & 37.5 & 48.8 & 61.3 \\
\ours (w/ all) & \textbf{26.8} & \textbf{41.4} & \textbf{51.9} & \textbf{62.8} \\
\bottomrule
\end{tabular}
\caption{Accuracy on WTQ test comparing \ours with baselines. Overall best results and best baseline results are in \textbf{bold} and \emph{italic} separately.}
\label{tab:exp_overall}
\end{table}

\begin{figure}[tb]
\centering
\includegraphics[width=\columnwidth, clip, keepaspectratio]{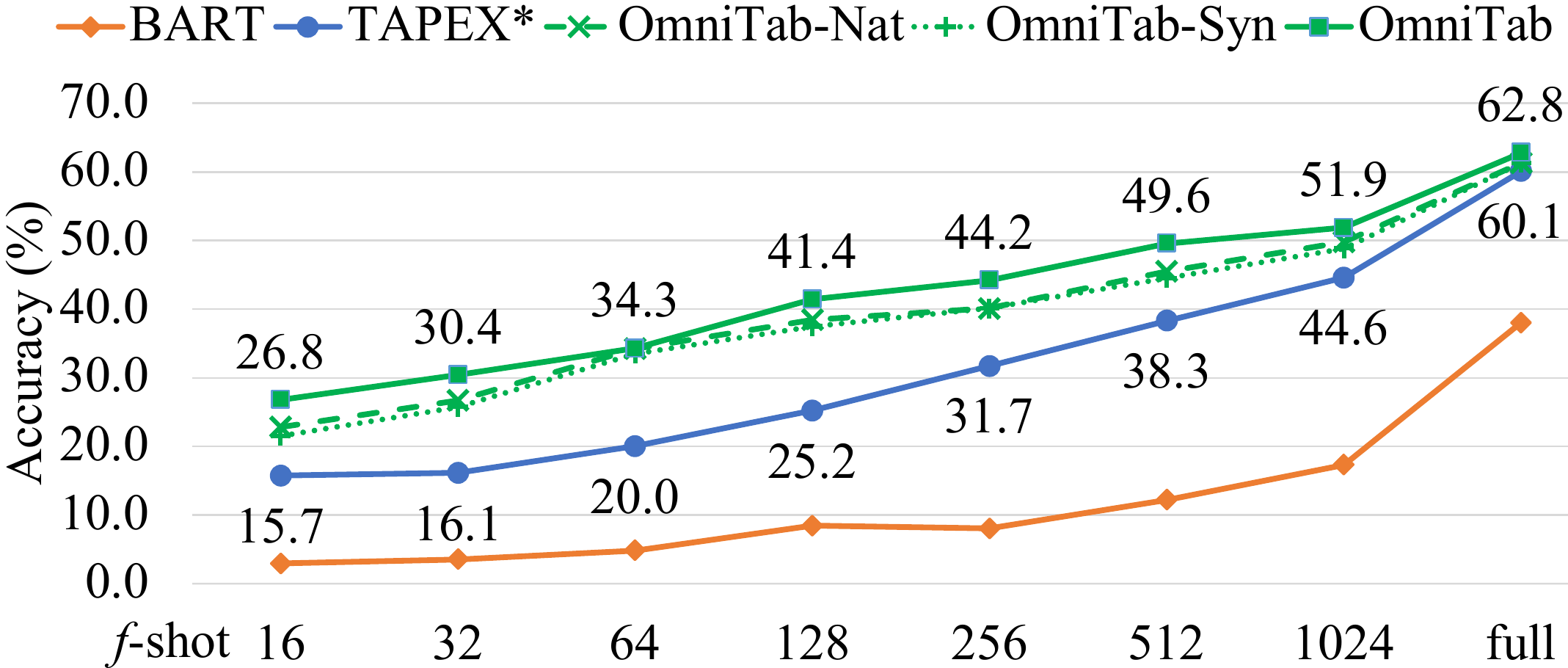}
\caption{WTQ test accuracy in all settings.}
\label{fig:annotation}
\end{figure}

\subsection{Overall Results}\label{sec:exp_overall}
The overall results comparing \ours with other baselines are listed in \autoref{tab:exp_overall}.
Across three few-shot settings, simulating low, medium, and high resource scenarios, pretraining on natural or synthetic data individually both outperform baselines significantly, and multitasking further increases the performance by a large margin.
\ours improves over the best baseline performance by 11.1\%, 16.2\%, and 6.4\% across the three settings, clearly demonstrating that pretraining on natural sentences relevant to tables and synthetic questions provides \ours with a stronger capacity to align text and tables and perform reasoning.
The two types of data are complementary to each other, which we will analyze in detail in \autoref{sec:exp_analysis}.
Despite the fact that we focus on the few-shot setting, we also observe significant improvements of 2.7\% on the full setting, establishing a new state-of-the-art on WTQ.
The performance in all few-shot/full settings shown in \autoref{fig:annotation} clearly indicates the superiority of \ours across the whole spectrum.
The improvement is larger when annotations are fewer, indicating the value of pretraining especially when fewer annotations are available.
We also observe improvements on \wikisql as shown in \autoref{tab:exp_wikisql}, reinforcing the effectiveness of our methods.

\begin{table}[tb]
\small
\centering
\begin{tabular}{lr@{\smallcol}r@{\smallcol}r@{\smallcol}r}
\toprule
\textbf{Model \hfill{$f$-shot:}} & \textbf{16} & \textbf{128} & \textbf{1024} & \textbf{full} \\
\midrule
TAPEX$^*$ & 43.4 & 63.6 & 75.6 & 88.1 \\
\ours & \textbf{63.6} & \textbf{75.6} & \textbf{82.9} & \textbf{88.7} \\
\bottomrule
\end{tabular}
\caption{Accuracy on \wikisql test.}
\label{tab:exp_wikisql}
\end{table}

\subsection{Ablation Study}\label{sec:exp_ablation}
Next, we study the contribution of individual components, including different retrieval methods, masking strategies, self-training methods, and varying the number of training pairs.

\paragraph{Comparison of Different Retrieval Methods}
Our first ablation concerns the influence of different retrieval methods on the final performance.
We examined three retrieval methods to pair tables with a relevant sentence, including string-based matching, BM25, and dense retrieval (\autoref{sec:natural}), as summarized in \autoref{tab:exp_retrieval}.
We also add a baseline (title-based heuristic) that pairs a table with the caption, article title, and description used by \citet{tapas-2020-herzig} to validate the utility of retrieval.
(1) Our three retrieval methods usually perform better than the title-based heuristic, indicating that retrieving sentences based on the table content is better than fixed heuristics that always pair a table with pre-specified content.
(2) By comparing two string-based matching variants, we found that selecting the sentence with the maximal number of mentions is better than sentences with minimal overlap,\footnote{Average \#mentions for max and min are 2.0 vs 1.2.} confirming our intuition that sentences more aligned with the table enable models to learn better alignment.
(3) BM25 performs similarly to string-based matching, partially because we still rely on string-based matching to locate mentions after BM25 returns a similar sentence.
(4) Dense retrieval with threshold $\tau=0.6$ achieves the best overall performance, but it is relatively sensitive to the threshold.
A high threshold only keeps highly relevant phrase-cell pairs, while a low threshold can discover more partial matches for masked pretraining, leading to a trade-off between quality and quantity.
Given that this initial attempt to use dense retrieval for text-table alignment directly uses BART without further tuning, further advances in retriever could likely improve this trade-off.

\begin{table}
\small
\centering
\begin{tabular}{l@{\smallcol}lr@{\smallcol}r@{\smallcol}r}
\toprule
\textbf{Model} & \hfill{\textbf{$f$-shot:}} & \textbf{16} & \textbf{128} & \textbf{1024} \\
\midrule
\multicolumn{2}{l}{TAPEX \cite{tapex-2021-liu}} & 10.4 & 23.1 & 45.5 \\
\midrule
\multicolumn{5}{c}{\emph{\ours w/ natural data obtained by}} \\
\multicolumn{2}{l}{title-based heuristic} & 21.5 & 34.2 & 48.4 \\
\emph{retrieval method} & \emph{other option} \\
\multirow{2}{*}{string-based} & (min) & 23.3 & 35.5 & 47.5 \\
 & (max) & \textbf{24.2} & 36.7 & 49.2 \\
\multicolumn{2}{l}{BM25} & 23.8 & 36.4 & 49.1 \\
\multirow{3}{*}{\underline{dense retrieval}} & ($\tau=0.5$) & 22.7 & 35.9 & 48.4 \\
 & ($\tau=0.7$) & 19.8 & 36.8 & 48.2 \\
 & \underline{($\tau=0.6$)} & 22.8 & \textbf{38.4} & \textbf{49.8} \\
\midrule
\multirow{2}{*}{\begin{tabular}{@{}c@{}} dense retrieval \\ ($\tau=0.6$) \\ \end{tabular}} & w/o salient mask & 17.5 & 33.6 & 47.1 \\
 & w/o random mask & 23.1 & 37.8 & 48.5 \\
\bottomrule
\end{tabular}
\caption{WTQ test accuracy when pretraining on natural data obtained by different retrieval methods, and using two masking strategies separately. Design choices used in our final model are \underline{underlined}.}
\label{tab:exp_retrieval}
\end{table}

\paragraph{Random Masking vs. Salient Masking}
We use both salient mention masking that only masks mentions of cells in the sentence and random masking in our final model.
To examine the contribution of each masking strategy, we remove one masking strategy from the underlined model at the bottom of \autoref{tab:exp_retrieval}.
It is clear that both maskings help, with salient masking being the major contributor, which indicates that masking tokens indicative of alignment is more effective than aimless masking.

\paragraph{Comparison of Different Self-training Methods}
To study which element is crucial in self-training, we perform ablations to study various aspects of self-training including (1) selection criterion for questions (generation- vs verification-based) and (2) models used for verification (BART vs \ours) by comparing all variants under the same setting of 128 annotated SQL-NL. 
As summarized in \autoref{tab:exp_selftraining}, self-training on selected questions with the highest generation probabilities given by the SQL2NL model does not improve over the baseline without self-training, which is mainly because the SQL2NL model is too weak to output reliable generation probabilities.
However, our method that selects questions with the highest probabilities to elicit answers from \ours (last line) improve over no self-training by a large margin (4.3\%, 2.5\%, and 1.8\%), validating the idea of leveraging the strong QA capacity of \ours to assess the quality of generated questions.
To confirm the source of success, we perform a sanity check that selects sentences most \emph{unlikely} to elicit answers (min), and the performance indeed becomes much lower.
We also replace \ours with BART that is only finetuned with 128 annotations, and the performance is significantly lower, confirming that stronger QA models can provide a better assessment.

\begin{table}
\small
\centering
\begin{tabular}{l@{\smallcol}l@{\smallcol}lr@{\smallcol}r@{\smallcol}r}
\toprule
\multicolumn{3}{l}{\textbf{Model} \hfill{\textbf{WTQ-}}} & \textbf{16} & \textbf{128} & \textbf{1024} \\
\midrule
\multicolumn{3}{l}{TAPEX \cite{tapex-2021-liu}} & 10.4 & 23.1 & 45.5 \\
\midrule
\multicolumn{6}{c}{\emph{\ours w/ synthetic data from SQL2NL trained}} \\
\emph{criteria} & \emph{op.} & \emph{verify}  \\
w/o self-training & - & - & 26.5 & 35.0 & 45.5 \\
w/ generation-based & max & - & 24.0 & 35.8 & 44.3 \\
\multirow{3}{*}{\underline{w/ verification-based}} & min & \ours & 15.5 & 27.4 & 41.9 \\
 & max & BART & 28.9 & 36.4 & 45.4 \\
 & \underline{max} & \underline{\ours} & \textbf{30.8} & \textbf{37.5} & \textbf{47.3} \\
\bottomrule
\end{tabular}
\caption{WTQ test accuracy when pretraining on synthetic data generated from an SQL2NL model trained with 128 annotations and various self-training methods. Design choices used in our final model are \underline{underlined}.}
\label{tab:exp_selftraining}
\end{table}

\begin{figure}[tb]
\centering
\includegraphics[width=\columnwidth, clip, keepaspectratio]{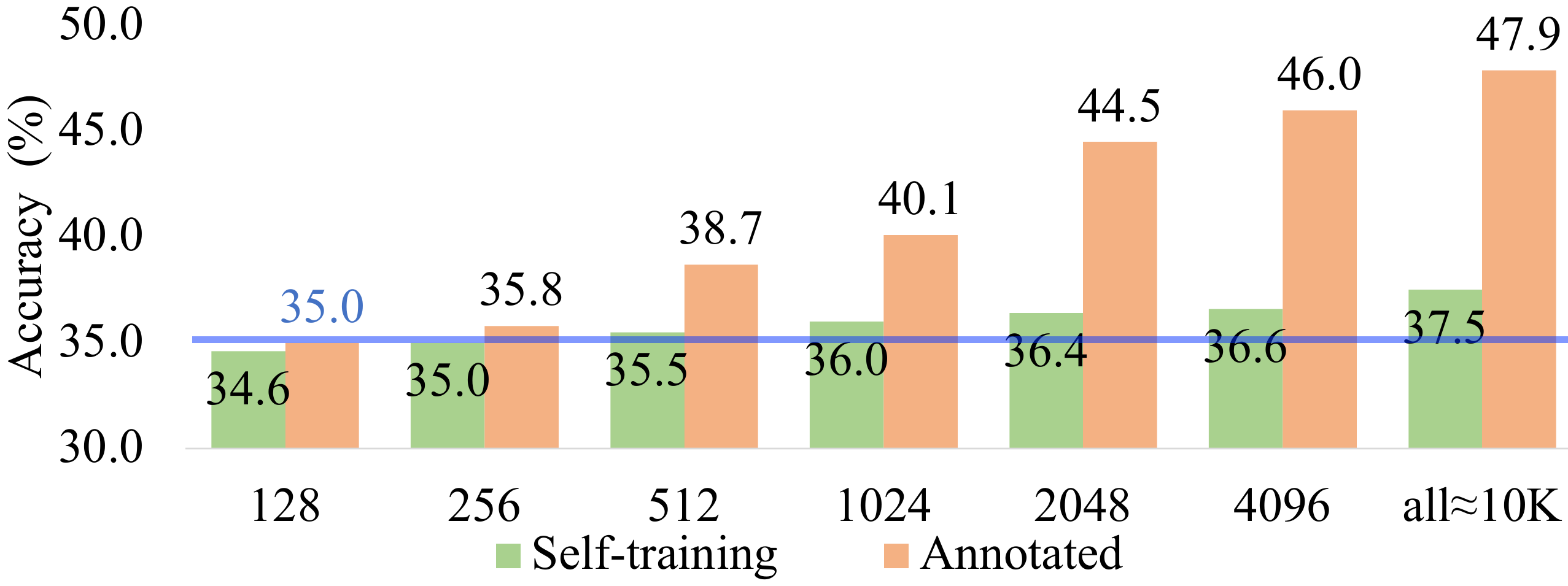}
\caption{WTQ test accuracy (128-shot) using different numbers of annotated and self-training SQL-NL pairs.}
\label{fig:exp_sql2nl}
\end{figure}

\paragraph{Performance w.r.t.~Number of Annotated and Self-training Pairs}
Here we study the influence of increasing either annotated or self-training SQL-NL pairs.
We use the SQL2NL model trained with 128 annotated pairs as a starting point, and additionally using more annotated or self-training pairs.
As shown in \autoref{fig:exp_sql2nl}, using more annotated or self-training pairs both improves over the initial performance of 35.0\%.
However, the improvement due to self-training still falls far behind the supervised approach, demonstrating the challenge of learning a robust SQL2NL model with very few annotations.
The increasing trend of self-training suggests that further improvements may be provided by using more pairs in self-training.

\begin{table}[tb]
\small
\centering
\begin{tabular}{@{}l@{\tinycol}c@{\smallcol}c@{}}
\toprule
\multicolumn{1}{r}{\textbf{Cases favoring:}} & \textbf{Natural} & \textbf{Synthetic} \\
\midrule
Avg \#tok in questions / SQL & 10.6 / 11.4 & 10.8 / \textbf{12.9} \\
Avg \#aligned question tok with tables & \textbf{1.8} & 1.0 \\
\bottomrule
\end{tabular}
\caption{Statistics of cases favoring natural vs synthetic data. Numbers indicating advantages are in \textbf{bold}.}
\label{tab:case_stat}
\end{table}

\begin{figure}[tb]
\centering
\includegraphics[width=\columnwidth, clip, keepaspectratio]{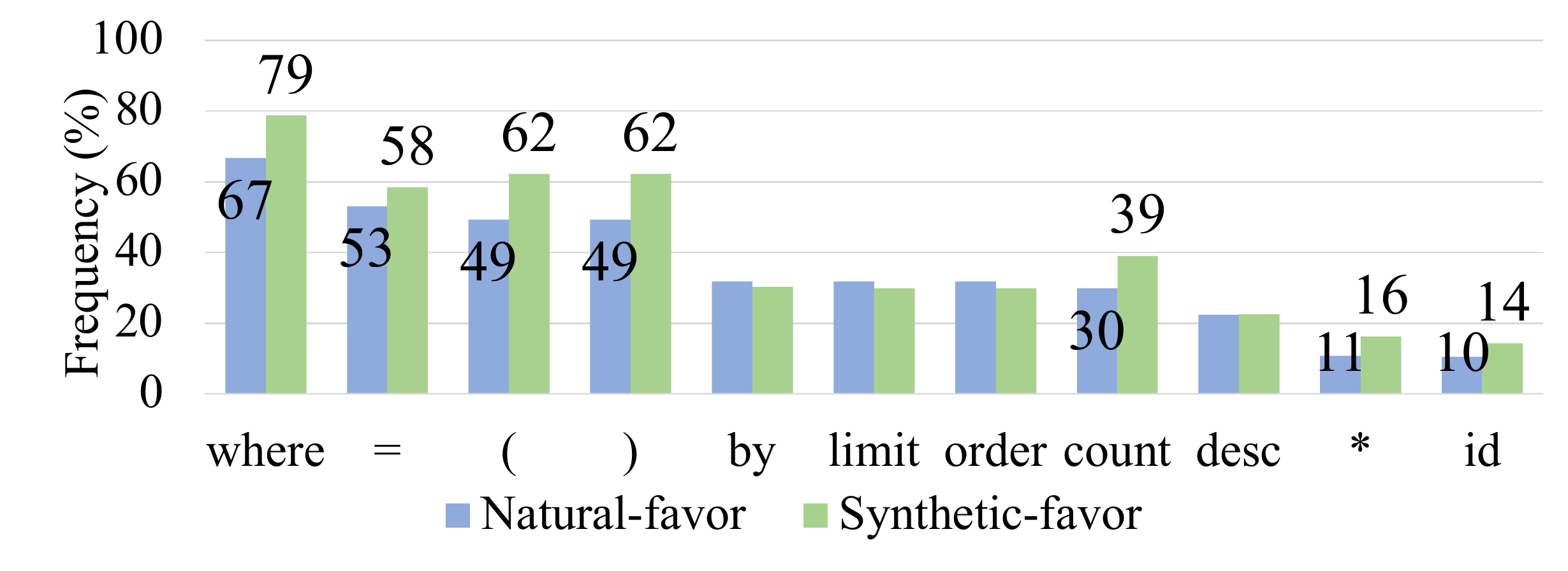}
\caption{Frequent SQL keywords in cases favoring natural vs synthetic data. Keywords with a large frequency difference are annotated with frequencies.}
\label{fig:case_sqlkw}
\end{figure}

\subsection{Analysis}\label{sec:exp_analysis}
\paragraph{Roles of Natural and Synthetic Data}
We quantitatively verified using the multitasking experiment that natural and synthetic are complementary to each other, with the hypothetical reason that natural data excels at enhancing alignment while synthetic data is more targeted on endowing reasoning capabilities.
Our analysis on cases where natural pretraining succeeds but synthetic fails and the opposite cases confirms that this is \emph{indeed} the case.
Enabled by the fine-grained annotation in \squall \cite{shi-2020-squall}, we compare \ours trained on natural or synthetic data separately in the 128-shot setting, and study on the two groups of cases in the development set of \wtq.
Based on 309/315 cases favoring natural/synthetic pretraining, we witness a clear distinction on the average number of question tokens aligned with tables between the two groups in \autoref{tab:case_stat}, indicating that natural data is indeed more targeted at addressing the alignment across formats.
We also compute the frequency of each SQL keyword for the two contrasting groups of cases. As shown in \autoref{fig:case_sqlkw}, cases favoring synthetic data indeed involves reasoning-rich keywords more frequently, such as ``where = ( )'' which are often used in nested queries, and ``count * id'' which are often used in aggregation.

\paragraph{Performance under Topical Distributional Shift}
Last, we analyze the robustness of \ours under topical distributional shift on WTQ-TS, which splits WTQ into five topics.
We finetune \ours on one topic (128-shot) and test the resulting model on all five topics.
As indicated by \autoref{tab:exp_topic}, \ours outperforms TAPEX by a large margin across all topics, validating the robustness of our methods under topical shift.

\begin{table}
\small
\centering
\begin{tabular}{lr@{\smallcol}r@{\smallcol}r@{\smallcol}r@{\smallcol}r}
\toprule
\textbf{Train/test} & \textbf{Sports} & \textbf{Culture} & \textbf{People} & \textbf{Politics} & \textbf{Misc} \\
\midrule
\textbf{Sports} & +16.3 & +14.1 & +15.8 & +14.4 & +18.8 \\
\textbf{Culture} & +13.9 & +12.7 & +14.3 & +13.0 & +10.3 \\
\textbf{People} & +21.5 & +14.6 & +20.6 & +14.6 & +17.5 \\
\textbf{Politics} & +18.5 & +14.3 & +17.8 & +16.0 & +13.5 \\
\textbf{Misc} & +18.3 & +14.7 & +17.2 & +14.6 & +14.9 \\
\bottomrule
\end{tabular}
\caption{Accuracy gain (128-shot) of \ours over TAPEX when finetuned on one topic and tested on all.}
\label{tab:exp_topic}
\end{table}

\section{Related Work}\label{sec:related}
Table-based QA is a well-studied area from early systems using structured queries as intermediate steps \cite{krishnamurthy-2017-typeparser,liang-2018-mapo,wang-2019-latent,tabert-2020-yin,yu-grappa-2021} to recent advances that generate answers in an end2end fashion \cite{tapas-2020-herzig,tapex-2021-liu}.
Our methods follow the end2end approach because of its modeling simplicity and higher flexibility.
Given large amounts of table and text on the web, table-based QA and other table-related tasks such as semantic parsing \cite{strug-2021-deng,shi-2021-gap} and table understanding \cite{turl-2020-deng,tuta-2021-wang} start witnessing efforts invested in pretraining on both structured and unstructured information.
Most works leveraging retrieval to find relevant information to assist pretraining are designed for text format only \cite{guu-2020-realm,lewis-2020-para}, while the majority of table-based pretraining still use alignment heuristics \cite{tapas-2020-herzig,tabert-2020-yin}.
There are some initial attempts to perform retrieval over tables \cite{ouz-2020-unik,herzig-2021-dtr,ma-2021-uni}, but they mainly use tables as an additional information source while we focus on pairing tables with text for pretraining.

\section{Conclusion}
We propose an omnivorous pretraining approach that consumes both natural and synthetic data to enhance the ability to understand and align text and tables and the ability to perform reasoning.
Our extensive results demonstrate the effectiveness of both data and verify their complementary value.
Our empirical results together with the case analysis indicate that omnivorous pretraining can indeed benefit from the merits of both data, encouraging future advances in retrieval and synthesis to obtain higher-quality data and better pretraining strategies to combine heterogeneous data.

\section*{Acknowledgments}
We thank Qian Liu and Pengcheng Yin for their insightful comments and suggestions.

\bibliography{custom}
\bibliographystyle{acl_natbib}

\end{document}